\documentclass[runningheads]{llncs}
\usepackage{graphicx,tabulary,multirow,xspace,xcolor}
\usepackage{comment}
\usepackage{amsmath,amssymb,soul,setspace,pifont}
\usepackage{makecell}
\usepackage{color}
\usepackage{lipsum}
\usepackage{xcolor}
\definecolor{darkgreen}{rgb}{0.13, 0.55, 0.13}

\usepackage[font=footnotesize]{caption}

\usepackage[algo2e,ruled,vlined]{algorithm2e}

\usepackage{breakcites}

\newcommand{\srunet}{SR-UNet\ }

\newcolumntype{x}[1]{>{\centering\arraybackslash}p{#1pt}}
\newlength\savewidth\newcommand\shline{\noalign{\global\savewidth\arrayrulewidth
  \global\arrayrulewidth 1pt}\hline\noalign{\global\arrayrulewidth\savewidth}}
\newcommand\hshline{\noalign{\global\savewidth\arrayrulewidth
  \global\arrayrulewidth 0.6pt}\hline\noalign{\global\arrayrulewidth\savewidth}}
\newcommand{\tablestyle}[2]{\setlength{\tabcolsep}{#1}\renewcommand{\arraystretch}{#2}\centering\footnotesize}
\makeatletter\renewcommand\paragraph{\@startsection{paragraph}{4}{\z@}
  {.5em \@plus1ex \@minus.2ex}{-.5em}{\normalfont\normalsize\bfseries}}\makeatother

\usepackage{booktabs}

\usepackage{algorithm}
\usepackage{listings}
\usepackage{etoolbox}
\makeatletter
\AfterEndEnvironment{algorithm}{\let\@algcomment\relax}
\AtEndEnvironment{algorithm}{\kern2pt\hrule\relax\vskip3pt\@algcomment}
\let\@algcomment\relax
\newcommand\algcomment[1]{\def\@algcomment{\footnotesize#1}}
\renewcommand\fs@ruled{\def\@fs@cfont{\bfseries}\let\@fs@capt\floatc@ruled
  \def\@fs@pre{\hrule height.8pt depth0pt \kern2pt}%
  \def\@fs@post{}%
  \def\@fs@mid{\kern2pt\hrule\kern2pt}%
  \let\@fs@iftopcapt\iftrue}
\makeatother

\begin{document}
\pagestyle{headings}
\mainmatter

\newcommand{\hide}[1]{}
\newcommand{\eg}{\emph{e.g.}\xspace}
\newcommand{\ie}{\emph{i.e.}\xspace}
\newcommand{\vs}{\emph{vs.}\xspace}

\newcommand\green[1]{\textcolor{darkgreen}{#1}}

\title{PointContrast: Unsupervised Pre-training for \\3D Point Cloud Understanding} 

\titlerunning{ECCV-20 submission ID \ECCVSubNumber} 
\authorrunning{ECCV-20 submission ID \ECCVSubNumber} 
\author{Saining Xie\inst{1} \and
Jiatao Gu\inst{1} \and 
Demi Guo\thanks{Work done while at Facebook AI Research.} \and 
Charles R. Qi$^{\star}$ \and \\
Leonidas Guibas\inst{2}$^{\star}$ \and 
Or Litany\inst{2}$^{\star}$
}
\authorrunning{S. Xie et al.}
\institute{Facebook AI Research \and
Stanford University
}
\titlerunning{PointContrast}
\maketitle
\begin{abstract}
   Arguably one of the top success stories of deep learning is transfer learning. The finding that pre-training a network on a rich source set (\eg, ImageNet) can help boost performance once fine-tuned on a usually much smaller target set, has been instrumental to many applications in language and vision. Yet, very little is known about its usefulness in 3D point cloud understanding. We see this as an opportunity considering the effort required for annotating data in 3D. In this work, we aim at facilitating research on 3D representation learning. Different from previous works, we focus on high-level scene understanding tasks. To this end, we select a suite of diverse datasets and tasks to measure the effect of unsupervised pre-training on a large source set of 3D scenes. Our findings are extremely encouraging: using a unified triplet of architecture, source dataset, and contrastive loss for pre-training, we achieve improvement over recent best results in segmentation and detection across 6 different benchmarks for indoor and outdoor, real and synthetic datasets -- demonstrating that the learned representation can generalize across domains. Furthermore, the improvement was similar to supervised pre-training, suggesting that future efforts should favor scaling data collection over more detailed annotation. We hope these findings will encourage more research on unsupervised pretext task design for 3D deep learning. Our code is publicly available at \url{https://github.com/facebookresearch/PointContrast}

\end{abstract}

\section{Introduction}
Representation learning is one of the main driving forces of deep learning research. In 2D vision, the finding that pre-training a network on a rich source set (\eg ImageNet classification) can help boost performance once fine-tuned on the usually much smaller target set, has been key to the success of many applications. A particularly important setting is when the pre-training stage is unsupervised, as this opens up the possibility to utilize a practically infinite train set size. Unsupervised pre-training has been remarkably successful in natural language processing \cite{radford2019language,devlin2018bert}, and has recently attracted increasing attention in 2D vision~\cite{oord2018representation,bachman2019learning,henaff2019data,tian2019contrastive,he2019momentum,oord2018representation,bachman2019learning,misra2019self,henaff2019data,wu2018unsupervised,hjelm2018learning,zhuang2019local,chen2020simple}. 

In the past few years, the field of 3D deep learning has witnessed much progress with an ever-increasing number of 3D representation learning schemes \cite{achlioptas2017learning,gadelha2018multiresolution,yang2018foldingnet,groueix2018papier,li2018so,wang2019deep,hassani2019unsupervised,elbaz20173d,zeng20173dmatch,deng2018ppf,choy20194d}. 
However, it still falls behind compared to its 2D counterpart as evidently, in all 3D scene understanding tasks, ad-hoc training \emph{from scratch} on the target data is still the dominant approach. Notably, all existing representation learning schemes are tested either on single objects or low-level tasks (e.g. registration). This status quo can be attributed to multiple reasons: 1) Lack of large-scale and high-quality data: compared to 2D images, 3D data is harder to collect, more expensive to label, and the variety of sensing devices may introduce drastic domain gaps; 2) Lack of unified backbone architectures: in contrast to 2D vision where architectures such as ResNets have proven successful as backbone networks for pre-training and fine-tuning, point cloud network architecture designs are still evolving; 3) Lack of a comprehensive set of datasets and high-level tasks for evaluation.

The purpose of this work is to move the needle by initiating research on \emph{unsupervised pre-training} with \emph{supervised fine-tuning} in deep learning for 3D scene understanding.  
To do so, we cover four important ingredients: 1) Selecting a large dataset to be used at pre-training; 2) identifying a backbone architecture that can be shared across many different tasks; 3) evaluating two unsupervised objectives for pre-training the backbone network, and 4) defining an evaluation protocol on a set of diverse downstream datasets and tasks.

Specifically, we choose ScanNet~\cite{dai2017scannet} as our source set on which the pre-training takes place, and utilize a sparse residual U-Net~\cite{ronneberger2015u,choy20194d} as the backbone architecture in all our experiments and focus on the point cloud representation of 3D data. For the pre-training objective, we evaluate two different contrastive losses: Hardest-contrastive loss~\cite{choy2019fully}, and PointInfoNCE -- an extension of InfoNCE loss~\cite{oord2018representation} used for pre-training in 2D vision. Next, we choose a broad set of target datasets and downstream tasks that includes: semantic segmentation on S3DIS~\cite{armeni_cvpr16}, ScanNetV2~\cite{dai2017scannet}, ShapeNetPart~\cite{Yi16} and Synthia 4D~\cite{ros2016synthia}; and object detection on SUN RGB-D~\cite{song2015sun,silberman2012indoor,janoch2013category,xiao2013sun3d} and ScanNetV2. Remarkably, our results indicate improved performance across all datasets and tasks (See Table~\ref{tab:dataset_summary} for a summary of the results). In addition, we found a relatively small advantage to pre-training with supervision. This implies that future efforts in collecting data for pre-training should favor scale over precise annotations.

Our contributions can be summarized as follows:
\begin{itemize}
    \item We evaluate, for the first time, the transferability of learned representation in 3D point clouds to high-level scene understanding.
    \item Our results indicate that \textit{unsupervised pre-training} improves performance across downstream tasks and datasets, while using a single unified architecture, source set and objective function.  
    \item Powered by unsupervised pre-training, we achieve a new state-of-the-art performance on 6 different benchmarks. 
    \item We believe these findings would encourage a change of paradigm on how we tackle 3D recognition and drive more research on 3D representation learning.   
\end{itemize}

\section{Related work}
\paragraph{Representation learning in 3D}
Deep neural networks are notoriously data hungry. This renders the ability to transfer learned representations between datasets and tasks extremely powerful. In 2D vision it has led to a surge of interest in finding optimal pretext unsupervised tasks \cite{pathak2016context,zhang2016colorful,zhang2017split,doersch2015unsupervised,noroozi2016unsupervised,gidaris2018unsupervised,caron2018deep,oord2018representation,bachman2019learning,misra2019self,henaff2019data,wu2018unsupervised,hjelm2018learning,zhuang2019local,chen2020simple,choy2019fully}. We note that while many of these tasks are \textit{low-level} (\eg pixel or patch level reconstruction), they are evaluated based on their transferability to \textit{high-level} tasks such as object detection. Being much harder to annotate, 3D tasks are potentially the biggest beneficiaries of unsupervised- and transfer-learning. This was shown in several works on single object tasks like reconstruction, classification and part segmentation~\cite{achlioptas2017learning,gadelha2018multiresolution,yang2018foldingnet,groueix2018papier,li2018so,wang2019deep,hassani2019unsupervised,sauder2019self}. Yet, generally much less attention has been devoted to representation learning in 3D that extends beyond the single-object level. Further, in the few cases that did study it, the focus was on low-level tasks like registration \cite{elbaz20173d,zeng20173dmatch,deng2018ppf}. In contrast, here we wish to push forward research in 3D representation learning by focusing on transferability to more high-level tasks on more complex scenes. 

\paragraph{Deep architectures for point cloud processing}
In this work, we focus on learning useful representation for point cloud data. Inspired by the success in 2D domain, we conjecture that an important ingredient in enabling such progress is the evident standardization of neural architectures. Canonical examples include VGGNet~\cite{simonyan2014very} and ResNet/ResNeXt~\cite{he2016deep,xie2017aggregated}. In contrast, point cloud neural network design is much less mature, as is apparent by the abundance of new architectures that have been recently proposed. This has multiple reasons. First, is the challenge of processing unordered sets \cite{qi2017pointnet,ravanbakhsh2016deep,zaheer2017deep,maron2020learning}. Second, is the choice of neighborhood aggregation mechanism which could either be hierarchical \cite{qi2017pointnetplusplus,klokov2017escape,zeng20183dcontextnet,gadelha2018multiresolution,lei2018spherical}, spatial CNN-like \cite{hua2018pointwise,xu2018spidercnn,li2018pointcnn,zhang2019shellnet,su2018splatnet}, spectral \cite{yi2017syncspeccnn,te2018rgcnn,wang2018local} or graph-based \cite{xie2018attentional,verma2018feastnet,wang2018dynamic,shen2018mining}. Finally, since the points are discrete samples of an underlying surface, continuous convolutions have also been considered \cite{wang2018deep,boulch2020convpoint,yang2020continuous}. 
Recently Choy et al. proposed the Minkowski Engine \cite{choy20194d}, an extension of submanifold sparse convolutional networks~\cite{graham20183d} to higher dimensions. In particular, sparse convolutional networks facilitate the adoption of common deep architectures from 2D vision, which in turn can help standardize deep learning for point cloud. In this work, we use a unified U-Net \cite{ronneberger2015u} architecture built with Minkowski Engine as the backbone network in all experiments and show it can gracefully transfer between tasks and datasets.

\section{PointContrast Pre-training}
In this section, we introduce our unsupervised pre-training pipeline. First, to motivate the necessity of a new pre-training scheme, we conduct a pilot study to understand the limitations of existing practice (pre-training on ShapeNet) in 3D deep learning (Section~\ref{sec.pilot}). After briefly reviewing an inspirational local feature learning work \emph{Fully Convolutional Geometric Features (FCGF)} (Section~\ref{sec.fcgf}), we introduce our unsupervised pre-training solution, \emph{PointContrast}, in terms of pretext task (Section~\ref{sec.pointcontrast}), loss function (Section~\ref{sec.loss}), network architecture (Section~\ref{sec.architecture}) and pre-training dataset (Section~\ref{sec.corpus}).
\subsection{Pilot Study: is Pre-training on ShapeNet Useful?}
\label{sec.pilot}
\begin{figure}[t]
\centering
\includegraphics[width=0.9\textwidth]{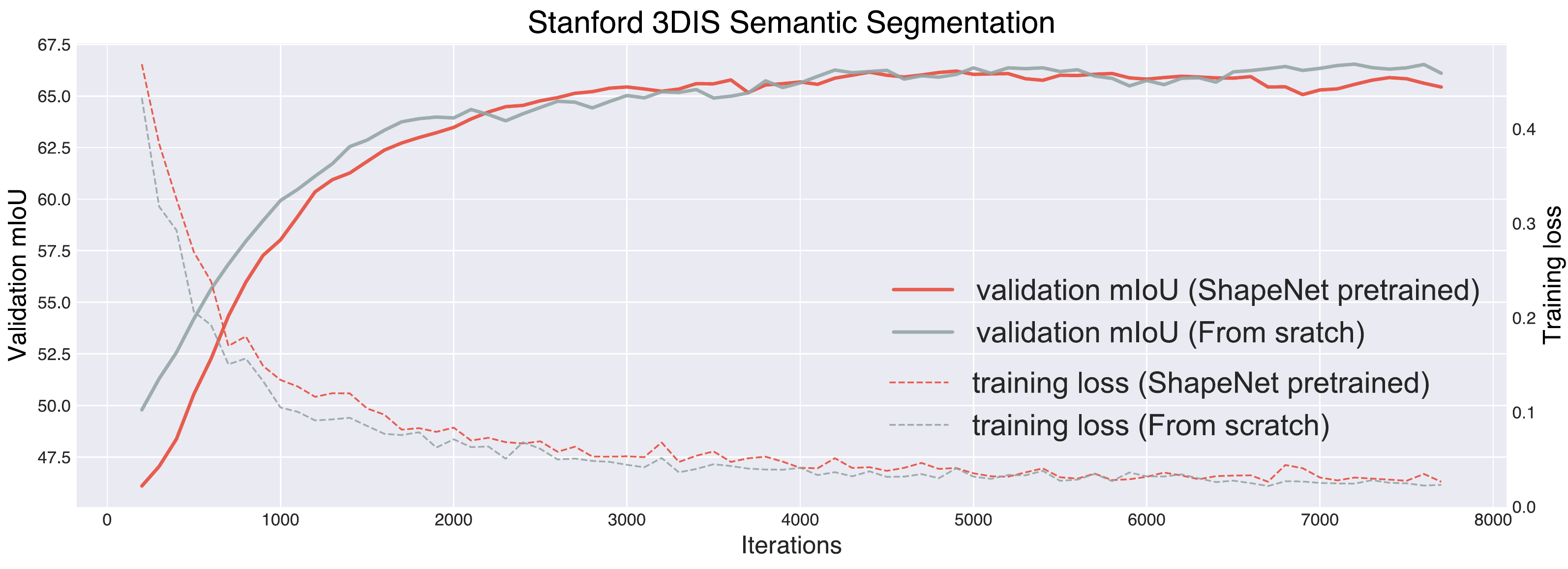}
\caption{\label{fig.3dis_shapenet} Training from scratch \vs fine-tuning with ShapeNet pre-trained weights.}
\end{figure}
Previous works on unsupervised 3D representation learning ~\cite{achlioptas2017learning,gadelha2018multiresolution,yang2018foldingnet,groueix2018papier,li2018so,wang2019deep,hassani2019unsupervised,sauder2019self} mainly focused on ShapeNet~\cite{shapenet2015}, a dataset of single-object CAD models. One underlying assumption is that by adopting ShapeNet as the ImageNet counterpart in 3D, features learned on \emph{synthetic single objects} could transfer to other real-world applications.
Here we take a step back and reassess this assumption by studying a straightforward supervised pre-training setup: we simply pre-train an encoder network on ShapeNet with \emph{full supervision}, and fine-tune it with a U-Net on a downstream task (S3DIS semantic segmentation). Following the practice in 2D representation learning, we use full supervision here as an upper bound to what could be gained from pre-training. We train a sparse ResNet-34 model (details to follow in Section~\ref{sec.architecture}) for 200 epochs. The model achieves a high validation accuracy of 85.4\% on ShapeNet classification task. In Figure~\ref{fig.3dis_shapenet}, we show the downstream task training curves for (a) training from scratch and (b) fine-tuning with ShapeNet pre-trained weights. Critically, one can observe that ShapeNet pre-training, even in the supervised fashion, \emph{hampers} downstream task learning. Among many potential explanations, we highlight two major concerns: 
\begin{itemize}
    \item \textbf{Domain gap between source and target data:} 
    Objects in ShapeNet are synthetic, normalized in scale, aligned in pose, and lack scene context. This makes pre-training and fine-tuning data distributions drastically different.
    \item \textbf{Point-level representation matters:} 
    In 3D deep learning, the local geometric features, \eg those encoded by a point and its neighbors, have proven to be discriminative and critical for 3D tasks~\cite{qi2017pointnet,qi2017pointnetplusplus}. Directly training on \emph{object instances} to obtain a global representation might be insufficient. 
\end{itemize} 

This led us to rethink the problem: if the goal of pre-training is to boost performance across many real-world tasks, exploring pre-training strategies on single objects might offer limited potential. 
(1) To address the domain gap concern, it might be beneficial to directly pre-train the network on complex scenes with multiple objects, to better match the target distributions; (2) to capture point-level information, we need to design a pretext task and corresponding network architecture that is not only based on instance-level/global representations, but instead can capture dense/local features at the point level. 
\begin{figure}[t]
\centering
\caption{\textbf{PointContrast: Pretext task for 3D pre-training.}}
\label{fig:overview}
\includegraphics[width=1.0\textwidth]{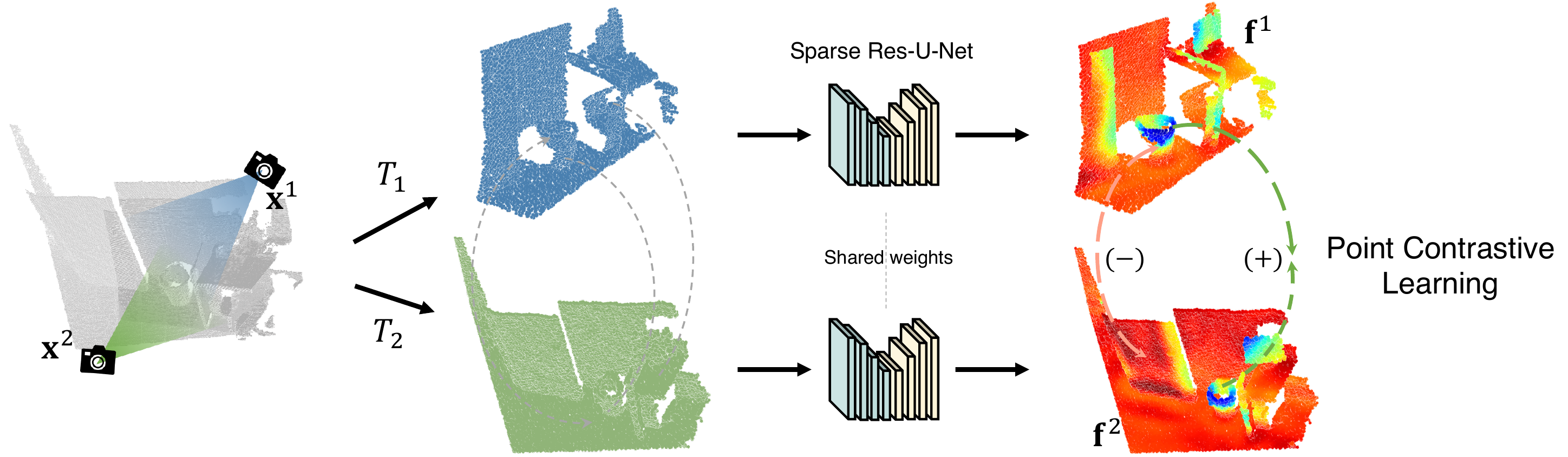}
\end{figure}
\begin{table}[t]
\tablestyle{4pt}{1.1}
\scalebox{0.9}{\begin{tabular}{lcccll}
\shline
\multicolumn{6}{c}{\textbf{PointContrast: Downstream Tasks for Fine-tuning}}\\
Datasets & \makecell{Real / \\Synth.} & Complexity & Env. & Task & \makecell{Rel.\\gain} \\
\shline
S3DIS & Real & \makecell{Entire floor,\\office} & Indoor & Segmentation & (\textcolor{darkgreen}{$+2.7\%$}) \scriptsize{mIoU}  \\
SUN RGB-D & Real & \makecell{Medium-sized\\ cluttered rooms} & Indoor & Detection & (\textcolor{darkgreen}{$+3.1\%$}) \scriptsize{mAP0.5} \\
\multirow{2}{*}{ScanNetV2} & \multirow{2}{*}{Real} & \multirow{2}{*}{Large rooms} & \multirow{2}{*}{Indoor} & Segmentation & (\textcolor{darkgreen}{$+1.9\%$}) \scriptsize{mIoU} \\
& & & & Detection & (\textcolor{darkgreen}{$+2.6\%$}) \scriptsize{mAP0.5} \\
ShapeNet & Synth. & Single objects & \makecell{Indoor \&\\outdoor} & Classification & (\textcolor{darkgreen}{$+4.0\%$}) \scriptsize{Acc.$^{\ast}$} \\
ShapeNetPart & Synth. & Object parts & \makecell{Indoor \&\\outdoor} & Segmentation & (\textcolor{darkgreen}{$+2.2\%$}) \scriptsize{mIoU$^{\ast}$} \\
Synthia 4D & Synth. & \makecell{Street scenes, \\ driving envs.} & Outdoor & Segmentation & (\textcolor{darkgreen}{$+3.3\%$}) \scriptsize{mIoU}\\
\shline
\end{tabular}}
\caption{\textbf{Summary of downstream fine-tuning tasks.} Compared to the baseline learning paradigm of training from scratch, which is dominant in 3D deep learning, our unsupervised pre-training method PointContrast boosts the performance across the board when finetuning on a diverse set of high-level 3D understanding tasks. \\$\ast$ indicates results trained using only $1\%$ of the training data.}
\label{tab:dataset_summary}
\end{table}

\subsection{Revisiting Fully Convolutional Geometric Features (FCGF)}
\label{sec.fcgf}
Here we revisit a previous approach 
FCGF~\cite{choy2019fully} designed to learn geometric features for \emph{low-level} tasks (\eg registration) as our work is mainly inspired by FCGF. FCGF is a deep learning based algorithm that learns local feature descriptors on correspondence datasets via metric learning. FCGF has two major ingredients that help it stand out and achieve impressive results in registration recall: (1) \textbf{a fully-convolutional design} and (2) \textbf{point-level metric learning}. With a fully-convolutional network (FCN)~\cite{long2015fully} design, FCGF operates on the entire input point cloud  (\eg full indoor or outdoor scenes) without having to crop the scene into patches as done in previous works; this way the local descriptors can aggregate information from a large number of neighboring points (up to the extent of receptive field size). As a result, point-level metric learning becomes natural. FCGF uses a U-Net architecture that has a full-resolution output (\ie for $N$ points, the network outputs $N$ associated feature vectors), and positive/negative pairs for metric learning are defined at the point level.

Despite having a fundamentally different goal in mind, FCGF offers inspirations that might address the pretext task design challenges: A fully-convolutional design will allow us to pre-train on the target data distributions that involve complex scenes with a large number of points, and we could define the pretext task directly on points. Under this perspective, we pose the question:
\textit{Can we repurpose FCGF as the pretext task for high-level 3D understanding?}

\subsection{PointContrast as a Pretext Task}
\label{sec.pointcontrast}
\begin{algorithm}[t]
\scriptsize
\SetAlgoLined
\textbf{Input:} Backbone architecture NN; Dataset $X=\{\mathbf{x}_i\in \mathbb{R}^{N\times 3}\}$; Point feature dimension $D$; \\
\textbf{Output:} Pre-trained weights for NN.\\
\For{\text{each point cloud} $\mathbf{x}$ \text{in} $X$}{
- From $\mathbf{x}$, generate two views $\mathbf{x}^1$ and $\mathbf{x}^2$.\\ 
- Compute correspondence mapping (matches) $M$ between points in $\mathbf{x}^1$ and $\mathbf{x}^2$.\\
- Sample two transformations $\mathbf{T}_1$ and $\mathbf{T}_2$.\\
- Compute point features $\mathbf{f}^1, \mathbf{f}^2 \in \mathbb{R}^{N\times D}$ by \\ $\mathbf{f}^1 = \text{NN}(\mathbf{T}_1(\mathbf{x}^1))$ and $\mathbf{f}^2 = \text{NN}(\mathbf{T}_2(\mathbf{x}^2))$.\\
- Backprop. to update NN with contrastive loss $\mathcal{L}_c(\mathbf{f}^1, \mathbf{f}^2)$ on the matched points.
}
\caption{General Framework of PointContrast}
\label{Alg1}
\end{algorithm}

FCGF focuses on local descriptor learning for low-level tasks only. In contrast, a good pretext task for pre-training aims to learn \emph{network weights} that are universally applicable and useful to many high-level 3D understanding tasks.
To take the inspiration of FCGF and create such pretext tasks, several design choices need to be revisited. In terms of \emph{architecture}, since inference speed is a major concern in registration tasks, the network used in FCGF is very light-weight; Contrarily, the success of pre-training relies on over-parameterized networks, as clearly evidenced in other domains~\cite{devlin2018bert,chen2020simple}. In terms of \emph{dataset}, FCGF uses domain-specific registration datasets such as 3DMatch~\cite{zeng20173dmatch} and KITTI odometry~\cite{geiger2013vision}, which lack both scale and generality. Finally, in terms of \emph{loss design}, contrastive losses explored in FCGF are tailored for registration and it is interesting to explore other alternatives.

In Algorithm~\ref{Alg1}, we summarize the overall pretext task framework explored in this work. We name the framework \emph{PointContrast}, since the high-level strategy of this pretext task is, contrasting---at the point level---between two transformed point clouds. Conceptually, given a point cloud $\textbf{x}$ sampled from a certain distribution, we first generate two views $\textbf{x}^1$ and $\textbf{x}^2$ that 
are aligned in the same world coordinates. We then compute the correspondence mapping $M$ between these two views. If $(i,j)\in M$ then point $\textbf{x}_i^1$ and point $\textbf{x}_j^2$ are a pair of matched points across two views. We then sample two random geometric transformations $T_1$ and $T_2$ to further transform the point clouds into two views. The transformation is what could make the pretext task challenging as the network needs to learn certain \emph{equivariance} to the geometric transformation imposed. 
In this work, we mainly consider rigid transformation including rotation, translation and scaling. Further details are provided in Appendix. Finally, a contrastive loss is defined over points in two views: we minimize the distance for matched points and maximize the distance of unmatched points. This framework, though coming from a very different motivation (metric learning for geometric local descriptors), shares a strikingly similar pipeline with recent contrastive-based methods for 2D unsupervised visual representation learning~\cite{wu2018unsupervised,he2019momentum,chen2020simple}. The key difference is that most work for 2D focuses on contrasting instances/images, while in our work the contrastive learning is done densely at the point level.

\subsection{Contrastive Learning Loss Design}
\label{sec.loss}
\subsubsection{Hardest-Contrastive Loss}
The first loss function, hardest-contrastive loss we try, is borrowed from the best-performing loss design proposed in FCGF~\cite{choy2019fully}, which adopts a hard negative mining scheme in traditional margin-based contrastive learning formulation,
{\tiny
\begin{align*}
 \mathcal{L}_c = \sum_{(i,j) \in \mathcal{P}} \bigg\{ \big[ d(\mathbf{f}_i,\mathbf{f}_j) - m_p \big]^2_+ \slash |\mathcal{P}| + 0.5 \big[m_n - \min_{k \in \mathcal{N}}d(\mathbf{f}_i,\mathbf{f}_k)\big]^2_+ \slash |\mathcal{N}_i| + 0.5\big[m_n - \min_{k \in \mathcal{N}}d(\mathbf{f}_j,\mathbf{f}_k)\big]^2_+ \slash |\mathcal{N}_j|
    \bigg\}
\end{align*}
}%
Here $\mathcal{P}$ is a set of matched (positive) pairs of points $\textbf{x}_i^1$ and $\textbf{x}_j^2$ from two views $\textbf{x}^1$ and $\textbf{x}^2$, and $\textbf{f}_i^1$ and $\textbf{f}_j^2$ are associated point features for the matched pair. $\mathcal{N}$ is a randomly sampled set of non-matched (negative) points which is used for the hardest negative mining, where the hardest sample is defined as the closest point in the $\mathcal{L}_2$ normalized feature space to a positive pair. $[x]_+$ denotes function $\max(0, x)$. $m_p = 0.1$ and $m_n = 1.4$ are margins for positive and negative pairs.
\subsubsection{PointInfoNCE Loss} Here we propose an alternative loss design for PointContrast. InfoNCE proposed in~\cite{oord2018representation} is widely used in recent unsupervised representation learning approaches for 2D visual understanding. By modeling the contrastive learning framework as a dictionary look-up process~\cite{he2019momentum}, InfoNCE poses contrastive learning as a classification problem and is implemented with a Softmax loss. Specifically, the loss encourages a query $q$ to be similar to its positive key $k^+$ and dissimilar to, typically many, negative keys $k^-$. 
One challenge in 2D is to scale the number of negative keys~\cite{he2019momentum}.

However, in the domain of 3D, we have a different problem: usually the real-world 3D datasets are much smaller in terms of instance count, but the number of points for each instance (\eg a indoor or outdoor scene) can be huge, \ie $100$K+ points even from one RGB-D frame. This unique property of 3D data property, together with the original motivation to modelling point level information, inspire us to propose the following PointInfoNCE loss:
{\footnotesize
\begin{align*}
\footnotesize
    \mathcal{L}_c = -\sum_{(i,j) \in \mathcal{P}}\;\;\;\log\frac{\exp(\mathbf{f}_i\cdot\mathbf{f}_j/\tau)}{\sum_{(\cdot, k) \in \mathcal{P}}\exp(\mathbf{f}_i\cdot\mathbf{f}_k/\tau)}
\end{align*}
}%
Here $\mathcal{P}$ is the set of all the positive matches from two views. In this formulation, we only consider points that have at least one match and do not use additional non-matched points as negatives. For a matched pair $(i,j) \in \mathcal{P}$, point feature $\textbf{f}^1_i$ will serve as the query and $\textbf{f}^2_j$ will serve as the positive key $k^+$. We use point feature $\textbf{f}^2_k$ where $\exists (\cdot, k) \in \mathcal{P}$ and $k \neq j$ as the set of negative keys. In practice, we sample a subset of 4096 matched pairs from $\mathcal{P}$ for faster training.

Compared to hardest-contrastive loss, the PointInfoNCE loss has a simpler formulation with fewer hyperparameters. Perhaps more importantly, due to a large number of negative distractors, it is more robust against \emph{mode collapsing} (features collapsed to a single vector) than the hardest-contrastive loss. In our experiments, we find that hard-contrastive loss is unstable and hard to train: the representation often collapses with extended training epochs (which is also observed in FCGF~\cite{choy2019fully}). 

\subsection{A Sparse Residual U-Net as Shared Backbone}
\label{sec.architecture}
We use a Sparse Residual U-Net (SR-UNet) architecture in this work. It is a 34-layer U-Net \cite{ronneberger2015u} architecture that has an encoder network of 21 convolution layers and a decoder network of 13 convolution/deconvolution layers. It follows the 2D ResNet basic block design and each conv/deconv layer in the network is followed by Batch Normalization (BN)~\cite{ioffe2015batch} and ReLU activation. The overall U-Net architecture has $37.85$M parameters. We provide more information and a visualization of the network in Appendix. The \srunet architecture was originally designed in \cite{choy20194d} that achieved significant improvement over prior methods on the challenging ScanNet semantic segmentation benchmark. In this work, we explore if we can use this architecture as a unified design for both the pre-training task and a diverse set of fine-tuning tasks. 

\subsection{Dataset for Pre-training}
\label{sec.corpus}
For local geometric feature learning approaches, including FCGF~\cite{choy2019fully}, training and evaluation are typically conducted on domain and task-specific datasets such as KITTI odometry~\cite{geiger2013vision} or 3DMatch~\cite{zeng20173dmatch}. Common registration datasets are typically constrained in either scale (training samples collected from just dozens of scenes), or generality (focusing on one specific application scenario, \eg indoor scenes or LiDAR scans for self-driving cars), or both. To facilitate future research on 3D unsupervised representation learning, in our work we utilize the ScanNet dataset for pre-training, aiming to address the scale issue. ScanNet is a collection of $\sim$1500 indoor scenes. Created with a light-weight RGB-D scanning procedure, ScanNet is currently the largest of its kind.\footnote{Admittedly, ScanNet is still much smaller in scale compared to 2D datasets.} 

Here we create a point cloud pair dataset on top of ScanNet for the pre-training framework shown in Figure~\ref{fig:overview}.  Given a scene $\textbf{x}$, we extract pairs of partial scans $\textbf{x}^1$ and $\textbf{x}^2$ from different views.
More precisely, for each scene, we first sub-sample RGB-D scans from the raw ScanNet videos every 25 frames, and align the 3D point clouds in the same world coordinates (by utilizing estimated camera poses for each frame). Then we collect point cloud pairs from the sampled frames and require that two point clouds in a pair have at least a $30\%$ overlap. We sample a total number of 870K point cloud pairs. Since the partial views are aligned in ScanNet scenes, it is straightforward to compute the correspondence mapping $M$ between two views with nearest neighbor search.

Although ScanNet only captures indoor data distributions, as we will see in Section~\ref{sec.synthia4d}, surprisingly it can generalize to other target distributions. We provide additional visualizations for the pre-training dataset in Appendix.

\section{Fine-tuning on Downstream Tasks}
The most important motivation for representation learning is to learn features that can transfer well to different downstream tasks. There could be different evaluation protocols to measure the usefulness of the learned representation. For example, probing with a linear classifier~\cite{goyal2019scaling}, or evaluating in a semi-supervised setup~\cite{henaff2019data}. The \emph{supervised fine-tuning} strategy, where the pre-trained weights are used as the initialization and are further refined on the target downstream task, is arguably the most practically meaningful way of evaluating feature transferability. with this setup, good features could directly lead to performance gains in downstream tasks. 

Under this perspective, in this section we perform extensive evaluations of the effectiveness of PointContrast framework by fine-tuning the pre-trained weights on multiple downstream tasks and datasets. We aim to cover a diverse suite of high-level 3D understanding tasks of different natures such as semantic segmentation, object detection and classification. In all experiment, we use the same backbone network, pre-trained on the proposed ScanNet pair dataset (Section~\ref{sec.corpus}) using both PointInfoNCE and Hardest-Constrastive objectives.
\subsection{ShapeNet: Classification and Part Segmentation}
\noindent\paragraph{Setup.} In Section~\ref{sec.pilot} we have observed that weights learned on supervised ShapeNet classification are not able to transfer well to scene-level tasks. Here we explore the opposite direction: Are PointContrast features learned on ScanNet useful for tasks on ShapeNet? To recap, ShapeNet \cite{shapenet2015} is a dataset of synthetic 3D objects of 55 common categories. It was curated by collecting CAD models from online open-sourced 3D repositories. In \cite{Yi16}, part annotations were added to a subset of ShapeNet models segmenting them into 2-5 parts. In order to provide a comparison with existing approaches, here we utilize the ShapeNetCore dataset (SHREC 15 split) for classification, and the ShapeNet part dataset for part segmentation, respectively. 
We uniformly sample point clouds of 1024 points from each model for classification and 2048 points for part segmentation. Albeit containing overlapping indoor object categories with ScanNet, this dataset is substantially different as it is synthetic and contains only single objects. 
We also follow recent works on 3D unsupervised representation learning~\cite{hassani2019unsupervised} to explore a more challenging setup: using a very small percentage (\eg 1\%-10\%) of training data to fine-tune the pre-trained model.

\noindent\paragraph{Results.} As shown in Table~\ref{tab:shapenet_classification} and Table~\ref{tab:shapenet_partseg}, for both datasets, the effectiveness of pre-training are correlated with the availability of training data. In the ShapeNet classification task (Table~\ref{tab:shapenet_classification}), pre-training helps most where less training data is available, achieving a $4.0\%$ improvement over the training-from-scratch baseline with the hardest-negative objective. We also note that ShapeNet is a class-imbalanced dataset and the minority (tail) classes are very infrequent. When using 100\% of the training data, pre-training provides a class-balancing effect, as it boosts performance more on underrepresented (tail) classes. 
Table~\ref{tab:shapenet_partseg} shows a similar effects of pre-training on part segmentation performance. Notably, using \srunet backbone architecture already boosts performance; yet, pre-training is able to provide further gains, especially when training data is scarce.  
\begin{table*}[t!]
\tablestyle{5pt}{1.1}
\scalebox{0.8}{
\begin{tabular}{l|l@{\hskip 0.5in} l@{\hskip 0.5in} l@{\hskip 0.2in}}
\textbf{evaluating on all 55 classes}  & 1\% data             & 10\% data           & 100\% data          \\ \hline
Trained from scratch     & 62.2          & 77.9          & 85.1          \\
PointConstrast (Hardest-Contrastive) & \textbf{66.2}~\tiny{\green{(+4.0)}} & \textbf{79.0}~\tiny{\green{(+1.1)}} & \textbf{85.7}~\tiny{\green{(+0.6)}} \\
PointConstrast (PointInfoNCE) & \textbf{65.8}~\tiny{\green{(+3.6)}} & \textbf{78.8}~\tiny{\green{(+0.9)}} & \textbf{85.7}~\tiny{\green{(+0.6)}} \\
\shline
\textbf{using 100\% training data}       &  10 tail classes     &  30 tail classes      & all 55 classes          \\ \hline
Train from scratch                 & 65.0          & 70.9          & 85.1          \\
PointConstrast (Hardest-Contrastive) & \textbf{70.9}~\tiny{\green{(+5.9)}} & \textbf{72.9}~\tiny{\green{(+2.0)}} & \textbf{85.7}~\tiny{\green{(+0.6)}} \\
PointConstrast (PointInfoNCE)        & \textbf{67.8}~\tiny{\green{(+2.8)}} & \textbf{72.0}~\tiny{\green{(+1.1)}} & \textbf{85.7}~\tiny{\green{(+0.6)}} \\
\end{tabular}}
\vspace{3mm}
\caption{\textbf{ShapeNet classification.} Top: classification accuracy with limited labeled training data for finetuning. Bottom: classification accuracy on the least represented classes in the data (tail-classes). 
In all cases, PointContrast boosts performance. Relative improvement increases with scarcer training data and on less frequent classes. 
}
\label{tab:shapenet_classification}
\end{table*}

\begin{table*}[t!]
\centering
\tablestyle{5pt}{1.1}
\scalebox{0.8}{\begin{tabular}{l|l@{\hskip 0.5in}l@{\hskip 0.5in}l@{\hskip 0.2in}}
 methods                 & IoU (1\% data)  & IoU (5\% data)           & IoU (100\% data)          \\ \hline
SO-Net\cite{li2018so}                  & 64.0 & 69.0 & -    \\
PointCapsNet\cite{zhao20193d}             & 67.0 & 70.0 & -    \\
Multitask Unsupervised\cite{hassani2019unsupervised}               & 68.2 & 77.7 & -    \\ \hline
Train from scratch  & 71.8 & 79.3 & 84.7 \\
PointConstrast (Hardest-Contrastive) & \textbf{74.0} \tiny{\green{(+2.2)}} & \textbf{79.9} \tiny{\green{(+0.6)}} & \textbf{85.1} \tiny{\green{(+0.4)}} \\
PointConstrast (PointInfoNCE)        & \textbf{73.1} \tiny{\green{(+1.3)}} & \textbf{79.9} \tiny{\green{(+0.6)}} & \textbf{85.1} \tiny{\green{(+0.4)}} \\
\end{tabular}}
\vspace{2mm}
\caption{\textbf{ShapeNet part segmentation.} Replacing the backbone architecture with \srunet already boosts performance. PointContrast pre-training further adds a significant gain, and outshines where labels are most limited. 
}
\label{tab:shapenet_partseg}
\end{table*}

\subsection{S3DIS Segmentation}
\noindent\paragraph{Setup.} Stanford Large-Scale 3D Indoor Spaces (S3DIS) \cite{armeni_cvpr16} dataset comprises 3D scans of 6 large-scale indoor areas collected from 3 office buildings. The scans are represented as point clouds and annotated with semantic labels of 13 object categories. Among the datasets used here for evaluation S3DIS is probably the most similar to ScanNet.
Transferring features to S3DIS represents a typical scenario for fine-tuning: the downstream task dataset is similar yet much smaller than the pre-training dataset. For the commonly used benchmark split (``Area 5 test''), there are only about 240 samples in the training set.
We follow \cite{choy20194d} for pre-processing, and use standard data augmentations. See Appendix for details.
\noindent\paragraph{Results.}
Results are summarized in Table~\ref{tab:S3DIS}. Again, merely switching the \srunet architecture, training from scratch already improves upon prior art. Yet, fine-tuning the features learned by PointContrast achieves markedly better segmentation results in mIoU and mAcc. Notably, the effect persists across both loss types, achieving a 2.7\% mIoU gain using Hardest-Contrastive loss and an on-par improvement of 2.1\% mIoU for the PointInfoNCE variant.
\begin{table*}[t!]
\centering
\tablestyle{1.2pt}{1.1}
\scalebox{0.8}{\begin{tabular}{l@{\hskip 1in} l@{\hskip 1in} l@{\hskip 0.5in}}
methods & mIoU & mAcc \\
\shline
PointNet~\cite{qi2017pointnet} & 41.1 & 49.0 \\
PointCNN~\cite{li2018pointcnn} & 57.3 & 63.9 \\
MinkowskiNet32~\cite{choy20194d} & 65.4 & 71.7 \\
\shline
Train from scratch & 68.2 & 75.5 \\
PointConstrast (Hardest-Contrastive) &  \textbf{70.9}~\tiny{\green{(+2.7)}} & \textbf{77.0}~\tiny{\green{(+1.5)}} \\
PointConstrast (PointInfoNCE) & 70.3~\tiny{\green{(+2.1)}} & 76.9~\tiny{\green{(+1.4)}}  \\
\end{tabular}}
\caption{\textbf{Stanford Area 5 Test (Fold 1) (S3DIS).} Replacing the backbone network with \srunet improves upon prior art. Using PointContrast adds further significant boost with a mild preference for Hardest-contrastive over the PointInfoNCE objective. See Appendix for more methods in comparison. 
}
\label{tab:S3DIS}
\end{table*}

\subsection{SUN RGB-D Detection} \label{sec:results:sunrgbd}
\noindent\paragraph{Setup.} We now attend to a different high-level 3D understanding task: object detection. Compared to segmentation tasks that estimate point labels, 3D object detection predicts 3D bounding boxes (localization) and their corresponding object labels (recognition). This calls for an architectural modification as the \srunet architecture does not directly output bounding box coordinates. Among many different choices \cite{yi2018gspn,hou20183d,qi2018frustum,imvotenet}, we identify the recently proposed VoteNet~\cite{voteNet} as a good candidate for three main reasons. First, VoteNet is designed to work directly on point clouds with no additional input (e.g. images).
Second, VoteNet originally uses PointNet++~\cite{qi2017pointnetplusplus} as the backbone architecture for feature extraction. Replacing this with a \srunet requires a minimal modification, keeping the proposal pipeline intact. In particular, we reuse the same hyperparameters. Third, VoteNet is the current state-of-the-art method that uses geometric features only, rendering an improvement markedly useful. 
We evaluate the detection performance on the SUN RGB-D dataset \cite{song2015sun}, a collection of single view RGB-D images.
The train set contains 5K images annotated with amodal, 3D oriented bounding boxes for objects from 37 categories.
\noindent\paragraph{Results.} 


\begin{table*}[t!]
\centering
\tablestyle{2pt}{1.1}
\scalebox{0.8}{\begin{tabular}{l@{\hskip 0.8in} l@{\hskip 0.5in}l@{\hskip 0.5in} l@{\hskip 0.2in}}
 methods & input & mAP@0.5 & mAP@0.25 \\
 \hshline
VoteNet~\cite{voteNet} & Geo & - &  57.0\\
VoteNet~\cite{voteNet} & Geo+Height & 32.9  & 57.7\\
 \hshline
Train from scratch & Geo & 31.7 & 55.6 \\
PointContrast(Hardest-Contrastive) & Geo & 34.5~\tiny{\green{(+2.8)}} & \textbf{57.5}~\tiny{\green{(+1.9)}}\\
PointContrast(PointInfoNCE) & Geo & \textbf{34.8}~\tiny{\green{(+3.1)}} & \textbf{57.5}~\tiny{\green{(+1.9)}}\\
\end{tabular}}
\caption{\textbf{SUN RGB-D detection results.} PointContrast demonstrates a substantial boost compared to training from scratch. We observe a larger improvement in localization as manifested by the $\Delta$mAP being larger for @$0.5$ than @$0.25$. }
\label{tab:sunrgbd:detection}
\end{table*}
We summarize the results in Table~\ref{tab:sunrgbd:detection}. We find that by simply switching in the backbone network, our baseline result (training from scratch) with the \srunet architecture achieves worse results (-1.4\% mAP@0.25). This may be attributed to the fact that  VoteNet design and hyperparameter settings were tailored to its PointNet++ backbone. However, PointContrast gracefully closes the gap by showing a +3.1\% gain on mAP@0.5, which also sets a new state-of-the-art in this metric. The performance gain with a harder evaluation metric (mAP@0.5) suggests that PointContrast pre-training can greatly help localization.  

\subsection{Synthia4D Segmentation}
\label{sec.synthia4d}
\noindent\paragraph{Setup.} Synthia4D \cite{ros2016synthia} is a large synthetic dataset designed to facilitate the training of deep neural networks for visual inference in driving scenarios. Photo-realistic renderings are generated from a virtual city, allowing dense and precise annotations of 13 semantic classes, together with pixel-accurate depth. We follow the train/val/test split as prescribed by \cite{choy20194d} in the clean setting. In the context of this work, Synthia4D is especially interesting since it is probably the most distant from our pre-training set (outdoor v.s. indoor, synthetic v.s. real). We test the segmentation performance using 3D \srunet on a per-frame basis.

\noindent\paragraph{Results.} 
\begin{table*}[t!]
\centering
\tablestyle{1.2pt}{1.1}
\scalebox{0.8}{\begin{tabular}{l@{\hskip 1in} l@{\hskip 1in} l@{\hskip 0.5in}}
methods & mIoU & mAcc \\
\shline
MinkowskiNet32~\cite{choy20194d} & 78.7 & 91.5 \\
\hshline
Train from scratch & 79.8 & 91.5 \\
PointContrast (Hardest-Contrastive) & 82.6~\tiny{\green{(+2.8)}} & {\bf 93.7}~\tiny{\green{(+2.2)}} \\
PointContrast (PointInfoNCE) & {\bf 83.1}~\tiny{\green{(+3.3)}} & {\bf 93.7}~\tiny{\green{(+2.2)}} \\

\end{tabular}}
\caption{\textbf{ Segmentation results on the 4D Synthia test set.} All networks here are \srunet with 3D kernels, trained on individual 3D frames without temporal modeling.}
\label{tab:synthia}
\end{table*} PointContrast pre-training brings substantial improvement over the baseline model trained from scratch (+2.3\% mIoU) as seen in Table~\ref{tab:synthia}. PointInfoNCE performs noticeably better than the hardest-contrastive loss. With unsupervised pre-training, the overall results are much better than the previous state-of-the-art reported in \cite{choy20194d}. Note that in \cite{choy20194d} it has been shown that adding the temporal learning (\ie using a 4D network instead of a 3D one) brings additional benefit. To use 3D pre-trained weights for a 4D network with an additional temporal dimension, we can simply inflate the convolutional kernels, following the standard practice in 2D video recognition \cite{carreira2017quo}. We leave it as future work.
\subsection{ScanNet: Segmentation and Detection}
\noindent\paragraph{Setup.} Although typically the source dataset for pre-training and the target dataset for fine-tuning are different, because of the specific multi-view contrastive learning pipeline for pre-training, PointContrast can likely learn different representations (\eg invariance/equivariance to rigid transformations or robustness to noise) compared to directly training with supervision. Thus it is interesting to see whether the pre-trained weights can further improve the results on ScanNet itself. We use ScanNet semantic segmentation and object detection tasks to test our hypothesis. For the segmentation experiment, we use the \srunet architecture to directly predict point labels. For the detection experiment, we again follow VoteNet~\cite{voteNet} and simply switch the original backbone network with the \srunet without other modifications to the detection head (See Appendix for details).
\noindent\paragraph{Results.} 
\begin{table*}[t!]
\centering
\tablestyle{1.2pt}{1.1}
\scalebox{0.8}{\begin{tabular}{l@{\hskip 1in}  l@{\hskip 1in} l@{\hskip 0.5in}}
methods & mIoU & mAcc \\
\shline
Train from scratch & 72.2 & 80.7 \\
PointContrast(Hardest-Contrastive) & 73.3~\tiny{\green{(+1.1)}} & 81.0~\tiny{\green{(+0.3)}} \\
PointContrast(PointInfoNCE) & \textbf{74.1}~\tiny{\green{(+1.9)}} & \textbf{81.6}~\tiny{\green{(+0.9)}} \\
\end{tabular}}
\caption{\textbf{ Segmentation results on ScanNet validation set.} PointContrast boosts performance on the ``in-domain'' transfer task where the pre-training and fine-tuning datasets come from a common source, showing the usefulness of pre-training even when labels are available.
}
\label{tab:scannet:segmentation}
\end{table*}

\begin{table*}[t!]
\centering
\tablestyle{1.2pt}{1.1}
\scalebox{0.8}{\begin{tabular}{l@{\hskip 0.5in} l@{\hskip 0.5in} l@{\hskip 0.5in} l@{\hskip 0.2in}}
 methods & input & mAP@0.5 & mAP@0.25 \\
\shline
DSS~\cite{song2016deep,hou20183d} & Geo+RGB         & 6.8  & 15.2 \\
3D-SIS~\cite{hou20183d} & Geo+RGB \tiny{(5 Views)}  & 22.5 & 40.2 \\
VoteNet~\cite{voteNet} & Geo+Height                 & 33.5 & 58.6 \\
\hshline
Train from scratch & Geo                         & 35.4 & 56.7 \\
PointContrast(Hardest-Contrastive) & Geo            & 37.3~\tiny{\green{(+1.9)}} & \textbf{59.2}~\tiny{\green{(+2.5)}}\\
PointContrast(PointInfoNCE) & Geo                   & \textbf{38.0}~\tiny{\green{(+2.6)}} & 58.5~\tiny{\green{(+1.8)}} \\
\end{tabular}}
\caption{\textbf{3D object detection results on ScanNet validation set.} Similarly to in-domain \textit{segmentation} task, here as well PointContrast boost performance on \textit{detection}, setting a new best result over prior art. See Appendix for more methods in comparison. 
}
\label{tab:scannet:detection}
\end{table*}
Results are summarized in Table~\ref{tab:scannet:segmentation} and Table~\ref{tab:scannet:detection}. Remarkably, on both detection and segmentation benchmark, models pre-trained with PointContrast outperform those trained from scratch. Notably, PointInfoNCE objective performs better than the Hardest-contrastive one, achieving a relative improvement of +1.9\% in terms of segmentation mIoU and +2.6\% in terms of detection mAP@0.5. Similar to SUN RGB-D detection, here we also observe that PointContrast features help most for localization as indicated by the larger margin of improvement for mAP@0.5 than mAP@0.25.
\subsection{Analysis Experiments and Discussions}
In this section, we show additional experiments to provide more insights on our pre-training framework. We use S3DIS segmentation for the experiments below.
\noindent\paragraph{Supervised pre-training.}
While the focus of this work is unsupervised pre-training, a natural baseline is to compare against supervised pre-training. To this end, we use the training-from-scratch baseline for the segmentation task on ScanNetV2 and fine-tune the network on S3DIS. This yields an mIoU of 71.2\%, which is only 0.3\% better than PointContrast unsupervised pre-training. We deem this a very \emph{encouraging signal} that suggests that the gap between supervised and unsupervised representation learning in 3D has been mostly closed (\emph{cf.} years of effort in 2D). One might argue that this is  due to the limited quality and scale of ScanNet, but even at this scale the amount of labor involved in annotating thousands of rooms is large. The outcome of this complements the conclusion we had so far: not only should we put resources into creating large-scale 3D datasets for pre-training; but if facing a trade-off between scaling the data size and annotating it, we should favor the former.
\noindent\paragraph{Fine-tuning vs from-scratch under longer training schedule.}
A recent study in 2D vision \cite{he2019rethinking} suggests that simply by training from scratch for more epochs might close the gap from ImageNet pre-training. We conduct additional experiments to train the network from scratch with $2\times$ and $3\times$ schedules on S3DIS, relative to the $1\times$ schedule of our default setup (10K iterations with batch size 48). We found that validation mIoU does not improve with longer training. In fact, the model exhibits overfitting due to the small dataset size, achieving $66.7\%$ and $66.1\%$ mIoU at 20K and 30K iteration, respectively. This suggests that potentially many of the 3D datasets could fall into the ``breakdown regime''\cite{he2019rethinking} where network pre-training is essential for good performance.
\noindent\paragraph{Holistic scene as a single view for PointContrast.}
To show that the multi-view design in PointContrast is important, we try a different variant where instead of having partial views $\mathbf{x}^1$ and $\mathbf{x}^2$, we directly use the reconstructed point cloud $\mathbf{x}$ (a full scene in ScanNet) PointContrast. We still apply independent transformations $T_1$ and $T_2$ to the same $\mathbf{x}$.
We tried different variants and augmentations such as random cropping, point jittering, and dropout. We also tried different transformations for $T_1$ and $T_2$ of different degrees of freedom. However, with the best configuration we can get a validation mIoU on S3DIS of $68.35$, which is just slightly better than the training from scratch baseline of $68.17$. This suggests that the multi-view setup in PointContrast is critical. Potential reasons include: much more abundant and diverse training samples; natural noise due to camera instability as good regularization, as also observed in \cite{zeng20173dmatch}.

\section{Conclusions}
We have demonstrated an extensive evaluation of the transferability of learned representations in 3D point clouds to high-level 3D understanding tasks. With the help of our unsupervised pre-training framework PointContrast, we achieve state-of-the-art results across 6 different benchmarks and demonstrate that the learned representation can generalize across domains. We hope these findings will encourage more research on 3D representation learning.

\paragraph{Acknowledgements.}
The authors would like to thank Chris Choy for his help in setting up the segmentation experiments with MinkowskiEngine. O.L. and L.G. were supported in part by NSF grant IIS-1763268, a Vannevar Bush Faculty Fellowship, and a grant from the SAIL-Toyota Center for AI Research.

\bibliographystyle{splncs04}
\bibliography{egbib}

\begin{thebibliography}{10}
\providecommand{\url}[1]{\texttt{#1}}
\providecommand{\urlprefix}{URL }
\providecommand{\doi}[1]{https://doi.org/#1}

\bibitem{achlioptas2017learning}
Achlioptas, P., Diamanti, O., Mitliagkas, I., Guibas, L.: Learning
  representations and generative models for {3D} point clouds. arXiv preprint
  arXiv:1707.02392  (2017)

\bibitem{armeni_cvpr16}
Armeni, I., Sener, O., Zamir, A.R., Jiang, H., Brilakis, I., Fischer, M.,
  Savarese, S.: {3D} semantic parsing of large-scale indoor spaces. In: ICCV
  (2016)

\bibitem{bachman2019learning}
Bachman, P., Hjelm, R.D., Buchwalter, W.: Learning representations by
  maximizing mutual information across views. In: NeurIPS (2019)

\bibitem{boulch2020convpoint}
Boulch, A.: Convpoint: continuous convolutions for point cloud processing.
  Computers \& Graphics  (2020)

\bibitem{caron2018deep}
Caron, M., Bojanowski, P., Joulin, A., Douze, M.: Deep clustering for
  unsupervised learning of visual features. In: ECCV (2018)

\bibitem{carreira2017quo}
Carreira, J., Zisserman, A.: Quo vadis, action recognition? a new model and the
  kinetics dataset. In: CVPR (2017)

\bibitem{shapenet2015}
Chang, A.X., Funkhouser, T., Guibas, L., Hanrahan, P., Huang, Q., Li, Z.,
  Savarese, S., Savva, M., Song, S., Su, H., Xiao, J., Yi, L., Yu, F.:
  {ShapeNet: An Information-Rich 3D Model Repository}. arXiv preprint
  arXiv:1512.03012  (2015)

\bibitem{chen2020simple}
Chen, T., Kornblith, S., Norouzi, M., Hinton, G.: A simple framework for
  contrastive learning of visual representations. arXiv preprint
  arXiv:2002.05709  (2020)

\bibitem{choy20194d}
Choy, C., Gwak, J., Savarese, S.: {4D} spatio-temporal convnets: Minkowski
  convolutional neural networks. In: CVPR (2019)

\bibitem{choy2019fully}
Choy, C., Park, J., Koltun, V.: Fully convolutional geometric features. In:
  ICCV (2019)

\bibitem{dai2017scannet}
Dai, A., Chang, A.X., Savva, M., Halber, M., Funkhouser, T., Nie{\ss}ner, M.:
  Scannet: Richly-annotated {3D} reconstructions of indoor scenes. In: CVPR
  (2017)

\bibitem{deng2018ppf}
Deng, H., Birdal, T., Ilic, S.: Ppf-foldnet: Unsupervised learning of rotation
  invariant {3D} local descriptors. In: ECCV (2018)

\bibitem{devlin2018bert}
Devlin, J., Chang, M.W., Lee, K., Toutanova, K.: {BERT}: Pre-training of deep
  bidirectional transformers for language understanding. In: NAACL (2019)

\bibitem{doersch2015unsupervised}
Doersch, C., Gupta, A., Efros, A.A.: Unsupervised visual representation
  learning by context prediction. In: ICCV (2015)

\bibitem{elbaz20173d}
Elbaz, G., Avraham, T., Fischer, A.: {3D} point cloud registration for
  localization using a deep neural network auto-encoder. In: CVPR (2017)

\bibitem{gadelha2018multiresolution}
Gadelha, M., Wang, R., Maji, S.: Multiresolution tree networks for {3D} point
  cloud processing. In: ECCV (2018)

\bibitem{geiger2013vision}
Geiger, A., Lenz, P., Stiller, C., Urtasun, R.: Vision meets robotics: The
  kitti dataset. The International Journal of Robotics Research
  \textbf{32}(11),  1231--1237 (2013)

\bibitem{gidaris2018unsupervised}
Gidaris, S., Singh, P., Komodakis, N.: Unsupervised representation learning by
  predicting image rotations. In: ICLR (2018)

\bibitem{goyal2019scaling}
Goyal, P., Mahajan, D., Gupta, A., Misra, I.: Scaling and benchmarking
  self-supervised visual representation learning. In: ICCV (2019)

\bibitem{graham20183d}
Graham, B., Engelcke, M., van~der Maaten, L.: 3d semantic segmentation with
  submanifold sparse convolutional networks. In: CVPR (2018)

\bibitem{groueix2018papier}
Groueix, T., Fisher, M., Kim, V.G., Russell, B.C., Aubry, M.: A
  papier-m{\^a}ch{\'e} approach to learning {3D} surface generation. In: CVPR
  (2018)

\bibitem{hassani2019unsupervised}
Hassani, K., Haley, M.: Unsupervised multi-task feature learning on point
  clouds. In: ICCV (2019)

\bibitem{he2019momentum}
He, K., Fan, H., Wu, Y., Xie, S., Girshick, R.: Momentum contrast for
  unsupervised visual representation learning. In: CVPR (2020)

\bibitem{he2019rethinking}
He, K., Girshick, R., Doll{\'a}r, P.: Rethinking imagenet pre-training. In:
  ICCV (2019)

\bibitem{he2017mask}
He, K., Gkioxari, G., Doll{\'a}r, P., Girshick, R.: Mask {R-CNN}. In: ICCV
  (2017)

\bibitem{he2016deep}
He, K., Zhang, X., Ren, S., Sun, J.: Deep residual learning for image
  recognition. In: CVPR (2016)

\bibitem{henaff2019data}
H{\'e}naff, O.J., Razavi, A., Doersch, C., Eslami, S., Oord, A.v.d.:
  Data-efficient image recognition with contrastive predictive coding. arXiv
  preprint arXiv:1905.09272  (2019)

\bibitem{hjelm2018learning}
Hjelm, R.D., Fedorov, A., Lavoie-Marchildon, S., Grewal, K., Bachman, P.,
  Trischler, A., Bengio, Y.: Learning deep representations by mutual
  information estimation and maximization. ICLR  (2019)

\bibitem{hou20183d}
Hou, J., Dai, A., Nie{\ss}ner, M.: {3D-SIS}: {3D} semantic instance
  segmentation of {RGB-D} scans. In: CVPR (2019)

\bibitem{hua2018pointwise}
Hua, B.S., Tran, M.K., Yeung, S.K.: Pointwise convolutional neural networks.
  In: CVPR (2018)

\bibitem{ioffe2015batch}
Ioffe, S., Szegedy, C.: Batch normalization: Accelerating deep network training
  by reducing internal covariate shift. In: ICML (2015)

\bibitem{janoch2013category}
Janoch, A., Karayev, S., Jia, Y., Barron, J.T., Fritz, M., Saenko, K., Darrell,
  T.: A category-level 3d object dataset: Putting the kinect to work. In:
  Consumer depth cameras for computer vision (2013)

\bibitem{klokov2017escape}
Klokov, R., Lempitsky, V.: Escape from cells: Deep kd-networks for the
  recognition of 3d point cloud models. In: ICCV (2017)

\bibitem{landrieu2018large}
Landrieu, L., Simonovsky, M.: Large-scale point cloud semantic segmentation
  with superpoint graphs. In: CVPR (2018)

\bibitem{lei2018spherical}
Lei, H., Akhtar, N., Mian, A.: Spherical convolutional neural network for 3d
  point clouds. arXiv preprint arXiv:1805.07872  (2018)

\bibitem{li2018so}
Li, J., Chen, B.M., Hee~Lee, G.: So-net: Self-organizing network for point
  cloud analysis. In: CVPR (2018)

\bibitem{li2018pointcnn}
Li, Y., Bu, R., Sun, M., Wu, W., Di, X., Chen, B.: Pointcnn: Convolution on
  x-transformed points. In: NeurIPS (2018)

\bibitem{long2015fully}
Long, J., Shelhamer, E., Darrell, T.: Fully convolutional networks for semantic
  segmentation. In: CVPR (2015)

\bibitem{maron2020learning}
Maron, H., Litany, O., Chechik, G., Fetaya, E.: On learning sets of symmetric
  elements. arXiv preprint arXiv:2002.08599  (2020)

\bibitem{misra2019self}
Misra, I., van~der Maaten, L.: Self-supervised learning of pretext-invariant
  representations. In: CVPR (2020)

\bibitem{noroozi2016unsupervised}
Noroozi, M., Favaro, P.: Unsupervised learning of visual representations by
  solving jigsaw puzzles. In: ECCV (2016)

\bibitem{oord2018representation}
Oord, A.v.d., Li, Y., Vinyals, O.: Representation learning with contrastive
  predictive coding. arXiv preprint arXiv:1807.03748  (2018)

\bibitem{pathak2016context}
Pathak, D., Krahenbuhl, P., Donahue, J., Darrell, T., Efros, A.A.: Context
  encoders: Feature learning by inpainting. In: CVPR (2016)

\bibitem{imvotenet}
Qi, C.R., Chen, X., Litany, O., Guibas, L.J.: Imvotenet: Boosting {3D} object
  detection in point clouds with image votes. In: CVPR (2020)

\bibitem{voteNet}
Qi, C.R., Litany, O., He, K., Guibas, L.J.: Deep hough voting for 3d object
  detection in point clouds. ICCV  (2019)

\bibitem{qi2018frustum}
Qi, C.R., Liu, W., Wu, C., Su, H., Guibas, L.J.: Frustum pointnets for 3d
  object detection from rgb-d data. In: CVPR (2018)

\bibitem{qi2017pointnet}
Qi, C.R., Su, H., Mo, K., Guibas, L.J.: Pointnet: Deep learning on point sets
  for 3d classification and segmentation. CVPR  (2017)

\bibitem{qi2017pointnetplusplus}
Qi, C.R., Yi, L., Su, H., Guibas, L.J.: Pointnet++: Deep hierarchical feature
  learning on point sets in a metric space. NeurIPS  (2017)

\bibitem{radford2019language}
Radford, A., Wu, J., Child, R., Luan, D., Amodei, D., Sutskever, I.: Language
  models are unsupervised multitask learners. OpenAI Blog  \textbf{1}(8), ~9
  (2019)

\bibitem{ravanbakhsh2016deep}
Ravanbakhsh, S., Schneider, J., Poczos, B.: Deep learning with sets and point
  clouds. arXiv preprint arXiv:1611.04500  (2016)

\bibitem{ronneberger2015u}
Ronneberger, O., Fischer, P., Brox, T.: U-net: Convolutional networks for
  biomedical image segmentation. In: International Conference on Medical image
  computing and computer-assisted intervention. pp. 234--241. Springer (2015)

\bibitem{ros2016synthia}
Ros, G., Sellart, L., Materzynska, J., Vazquez, D., Lopez, A.M.: The synthia
  dataset: A large collection of synthetic images for semantic segmentation of
  urban scenes. In: CVPR (2016)

\bibitem{sauder2019self}
Sauder, J., Sievers, B.: Self-supervised deep learning on point clouds by
  reconstructing space. In: NeurIPS (2019)

\bibitem{shen2018mining}
Shen, Y., Feng, C., Yang, Y., Tian, D.: Mining point cloud local structures by
  kernel correlation and graph pooling. In: CVPR (2018)

\bibitem{silberman2012indoor}
Silberman, N., Hoiem, D., Kohli, P., Fergus, R.: Indoor segmentation and
  support inference from {RGB-D} images. ECCV  (2012)

\bibitem{simonyan2014very}
Simonyan, K., Zisserman, A.: Very deep convolutional networks for large-scale
  image recognition. arXiv preprint arXiv:1409.1556  (2014)

\bibitem{song2015sun}
Song, S., Lichtenberg, S.P., Xiao, J.: Sun rgb-d: A rgb-d scene understanding
  benchmark suite. In: CVPR (2015)

\bibitem{song2016deep}
Song, S., Xiao, J.: Deep sliding shapes for amodal 3d object detection in rgb-d
  images. In: CVPR (2016)

\bibitem{su2018splatnet}
Su, H., Jampani, V., Sun, D., Maji, S., Kalogerakis, E., Yang, M.H., Kautz, J.:
  Splatnet: Sparse lattice networks for point cloud processing. In: CVPR (2018)

\bibitem{tatarchenko2018tangent}
Tatarchenko, M., Park, J., Koltun, V., Zhou, Q.Y.: Tangent convolutions for
  dense prediction in 3d. In: CVPR (2018)

\bibitem{tchapmi2017segcloud}
Tchapmi, L., Choy, C., Armeni, I., Gwak, J., Savarese, S.: Segcloud: Semantic
  segmentation of 3d point clouds. In: 3DV (2017)

\bibitem{te2018rgcnn}
Te, G., Hu, W., Zheng, A., Guo, Z.: Rgcnn: Regularized graph cnn for point
  cloud segmentation. In: ACM Multimedia (2018)

\bibitem{tian2019contrastive}
Tian, Y., Krishnan, D., Isola, P.: Contrastive multiview coding. arXiv preprint
  arXiv:1906.05849  (2019)

\bibitem{verma2018feastnet}
Verma, N., Boyer, E., Verbeek, J.: Feastnet: Feature-steered graph convolutions
  for {3D} shape analysis. In: CVPR (2018)

\bibitem{wang2018local}
Wang, C., Samari, B., Siddiqi, K.: Local spectral graph convolution for point
  set feature learning. In: ECCV (2018)

\bibitem{wang2018deep}
Wang, S., Suo, S., Ma, W.C., Pokrovsky, A., Urtasun, R.: Deep parametric
  continuous convolutional neural networks. In: CVPR (2018)

\bibitem{wang2019deep}
Wang, Y., Solomon, J.M.: Deep closest point: Learning representations for point
  cloud registration. In: ICCV (2019)

\bibitem{wang2018dynamic}
Wang, Y., Sun, Y., Liu, Z., Sarma, S.E., Bronstein, M.M., Solomon, J.M.:
  Dynamic graph cnn for learning on point clouds. ACM TOG  \textbf{38}(5),
  1--12 (2019)

\bibitem{wu2018unsupervised}
Wu, Z., Xiong, Y., Yu, S.X., Lin, D.: Unsupervised feature learning via
  non-parametric instance discrimination. In: CVPR (2018)

\bibitem{xiao2013sun3d}
Xiao, J., Owens, A., Torralba, A.: {SUN3D}: A database of big spaces
  reconstructed using sfm and object labels. In: ICCV (2013)

\bibitem{xie2017aggregated}
Xie, S., Girshick, R., Doll{\'a}r, P., Tu, Z., He, K.: Aggregated residual
  transformations for deep neural networks. In: CVPR (2017)

\bibitem{xie2018attentional}
Xie, S., Liu, S., Chen, Z., Tu, Z.: Attentional shapecontextnet for point cloud
  recognition. In: CVPR (2018)

\bibitem{xu2018spidercnn}
Xu, Y., Fan, T., Xu, M., Zeng, L., Qiao, Y.: Spidercnn: Deep learning on point
  sets with parameterized convolutional filters. In: ECCV (2018)

\bibitem{yang2018foldingnet}
Yang, Y., Feng, C., Shen, Y., Tian, D.: Foldingnet: Point cloud auto-encoder
  via deep grid deformation. In: CVPR (2018)

\bibitem{yang2020continuous}
Yang, Z., Litany, O., Birdal, T., Sridhar, S., Guibas, L.: Continuous geodesic
  convolutions for learning on {3D} shapes. In: arXiv preprint arXiv:2002.02506
  (2020)

\bibitem{ye20183d}
Ye, X., Li, J., Huang, H., Du, L., Zhang, X.: 3d recurrent neural networks with
  context fusion for point cloud semantic segmentation. In: ECCV (2018)

\bibitem{Yi16}
Yi, L., Kim, V.G., Ceylan, D., Shen, I.C., Yan, M., Su, H., Lu, C., Huang, Q.,
  Sheffer, A., Guibas, L.: A scalable active framework for region annotation in
  {3D} shape collections. SIGGRAPH Asia  (2016)

\bibitem{yi2017syncspeccnn}
Yi, L., Su, H., Guo, X., Guibas, L.J.: Syncspeccnn: Synchronized spectral cnn
  for {3D} shape segmentation. In: CVPR (2017)

\bibitem{yi2018gspn}
Yi, L., Zhao, W., Wang, H., Sung, M., Guibas, L.: {GSPN}: Generative shape
  proposal network for {3D} instance segmentation in point cloud. In: CVPR
  (2019)

\bibitem{zaheer2017deep}
Zaheer, M., Kottur, S., Ravanbakhsh, S., Poczos, B., Salakhutdinov, R.R.,
  Smola, A.J.: Deep sets. In: NeurIPS (2017)

\bibitem{zeng20173dmatch}
Zeng, A., Song, S., Nie{\ss}ner, M., Fisher, M., Xiao, J., Funkhouser, T.:
  {3DMatch}: Learning local geometric descriptors from {RGB-D} reconstructions.
  In: CVPR (2017)

\bibitem{zeng20183dcontextnet}
Zeng, W., Gevers, T.: {3DContextNet}: {KD} tree guided hierarchical learning of
  point clouds using local and global contextual cues. In: ECCV (2018)

\bibitem{zhang2016colorful}
Zhang, R., Isola, P., Efros, A.A.: Colorful image colorization. In: ECCV (2016)

\bibitem{zhang2017split}
Zhang, R., Isola, P., Efros, A.A.: Split-brain autoencoders: Unsupervised
  learning by cross-channel prediction. In: CVPR (2017)

\bibitem{zhang2019shellnet}
Zhang, Z., Hua, B.S., Yeung, S.K.: Shellnet: Efficient point cloud
  convolutional neural networks using concentric shells statistics. In: ICCV
  (2019)

\bibitem{zhao20193d}
Zhao, Y., Birdal, T., Deng, H., Tombari, F.: {3D} point capsule networks. In:
  CVPR (2019)

\bibitem{zhuang2019local}
Zhuang, C., Zhai, A.L., Yamins, D.: Local aggregation for unsupervised learning
  of visual embeddings. In: ICCV (2019)

\end{thebibliography}

\newpage
\appendix
\section{Visualization of the SR-UNet Architecture}
\label{app:supp_sr-unet}
Here we show the SR-UNet architecture that is used as a shared backbone in our paper for both the pre-training and the fine-tuning phases. This U-Net architecture was originally proposed in \cite{choy20194d} for ScanNet semantic segmentation.
\begin{figure}[!htp]
\centering
\includegraphics[width=1.0\textwidth]{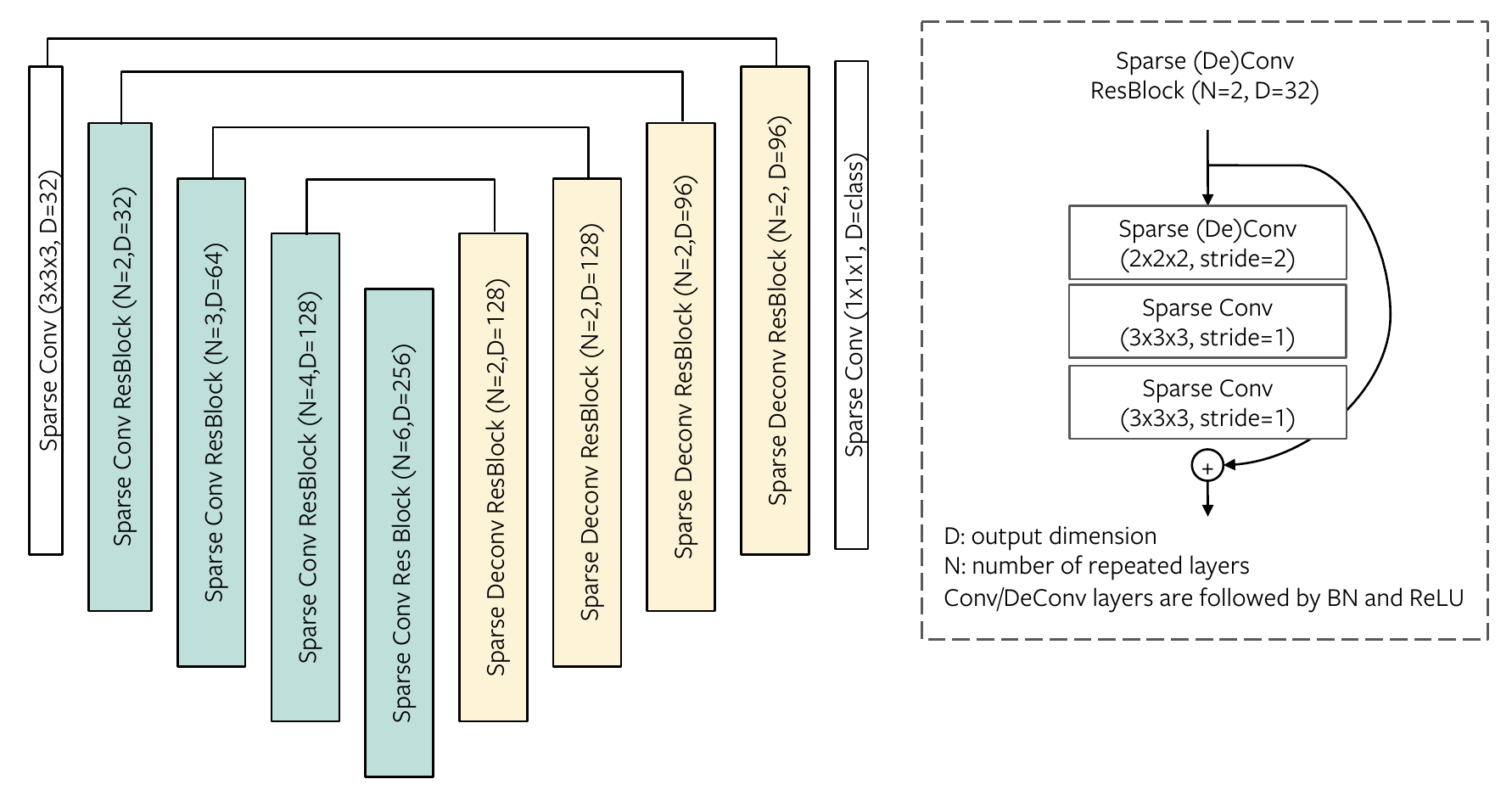}
\caption{\srunet architecture we used as a shared backbone network for pre-training and fine-tuning tasks. For segmentation and detection tasks, both the encoder and decoder weights are fine-tuned; for classification downstream tasks, only the encoder network is kept and fine-tuned.}
\label{fig:architecture}
\end{figure}

\section{Visualization of the ScanNet Point Cloud Pair Dataset}
\label{app:visualization_pc}
\begin{figure}[H]
\centering
\includegraphics[width=1.0\textwidth]{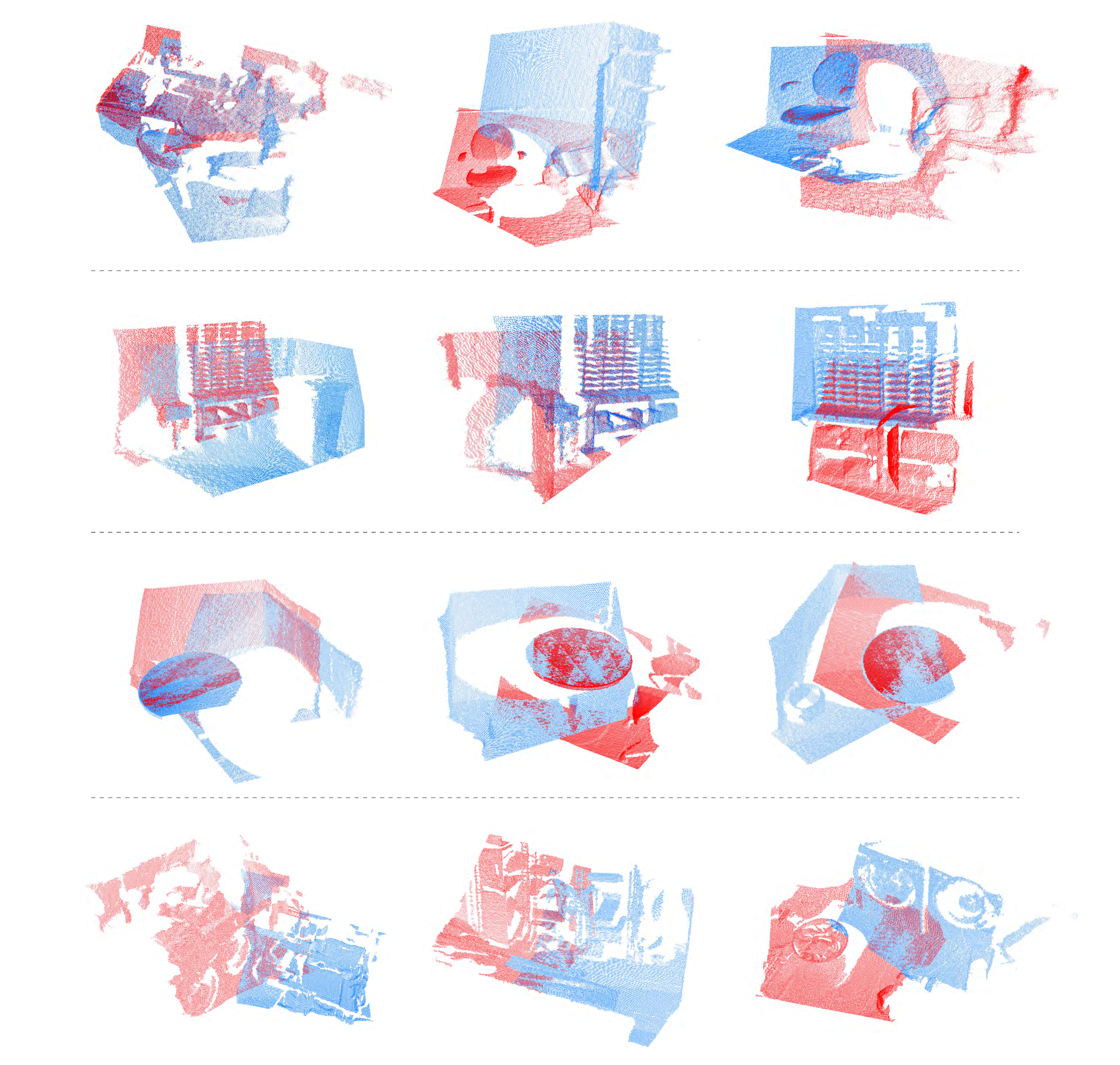}
\caption{Visualization of the ScanNet point cloud pair dataset used for pre-training. Each row is a randomly sampled scene. Each column is a different pair of point clouds sampled from the same scene. Different colors are corresponding to two different views (partial scans). At least 30\% of the points are overlapping in two views.}
\label{fig:visualization_pc}
\end{figure}

\section{ShapeNet Supervised Training Details}
We use a sparse ResNet network that has an identical structure to the encoder part of the SR-UNet in Appendix~\ref{app:supp_sr-unet}. We use Adam optimizer, and add standard data augmentations including rotation, scaling and translation, following \cite{qi2017pointnet,qi2017pointnetplusplus}. We perform a grid search over the learning rate, weight decay, voxel size (for sparse convolution), and the number of input points. 
The best performing model configuration is learning rate 0.004, voxel size 0.01, weight decay 1e-5, batch size 512 and 2048 input points. The 85.4\% accuracy is to our knowledge the best results that have been reported on this SHREC benchmark split. We use 8 Titan-V100 GPU with data parallelism to train the model. We train the model for 200 epochs and the training takes around 8 hours.

\section{Details on PointContrast Pre-training}
\subsection{Details on Transformations}
\label{app:transformation-details}
The transformations $\mathbf{T}_1$ and $\mathbf{T}_2$ applied to two views $\mathbf{x}^1$ and $\mathbf{x}^2$ in our experiments involves a random rotation ($0$ to $360^\circ$) along an arbitrary axis (applied independently to both views). We apply scale augmentation to both views (0.8$\times$ to 1.2$\times$ of the input scale). We have experimented with other augmentations such as translation, point coordinate jittering, and point dropout and did not find noticeable difference in fine-tuning performances.
\subsection{Details on Loss Functions}
For the hardest-contrastive loss, the positive sample size is 1024 and the hardest negative sample size is 256. More details can be found in \cite{choy2019fully}. For the PointInfoNCE loss, we provide a detailed PyTorch-like pseudo-code (and explanatory comments) in Algorithm~\ref{alg:code}. 
\begin{algorithm}[h]
\caption{Pseudocode of PointInfoNCE Loss implementation.}
\label{alg:code}
\algcomment{\fontsize{7.2pt}{0em}\selectfont \texttt{mm}: matrix multiplication;
}
\definecolor{codeblue}{rgb}{0.25,0.5,0.5}
\lstset{
  backgroundcolor=\color{white},
  basicstyle=\fontsize{7.2pt}{7.2pt}\ttfamily\selectfont,
  columns=fullflexible,
  breaklines=true,
  captionpos=b,
  commentstyle=\fontsize{7.2pt}{7.2pt}\color{codeblue},
  keywordstyle=\fontsize{7.2pt}{7.2pt},
}
\begin{lstlisting}[language=python]
# f_v1, f_v2: features for matched points (in a minibatch) between view 1 and view 2: NxC
# NN: shared backbone network
# t: temperature
# Ns: subsampling size for point features.

f_v1,inds = random.choice(f_v1, Ns, dim=0)  # subsample view 1 point features
f_v2 = f_v2[inds,:]  # subsample view 2 point features

logits = torch.mm(f_v1, f_v2.transpose(1, 0)) # Ns by Ns
labels = torch.arange(Ns) # for k-th row, the positive sample is at the k-th position.
loss = CrossEntropyLoss(logits/t, labels)

# SGD update: shared backbone network
loss.backward()
update(NN.params)
\end{lstlisting}
\end{algorithm}
\section{S3DIS Segmentation Experimental Details}
Here we provide training details for S3DIS semantic segmentation task. We use the widely adopted Area 5 Test (Fold 1) split for training and testing. For all the PointContrast variants (Training from scratch, Hardest-contrastive Pretrained, and PointInfoNCE Pretrained) we use the same hyperparameter settings. Specifically we train the model with 8 V100 GPUs with data parallelism for 10,000 iterations. Batch size is 48. Batch normalization is applied independently on each GPU. We use SGD+momentum optimizer with an initial learning rate 0.8. We use Polynomial LR scheduler with a power factor of 0.9. Weight decay is 0.0001 and voxel size is 0.05 (5cm). We use the same data augmentation techniques in \cite{choy20194d} such as color hue/saturation augmentation and jittering, as well as scale augmentations (0.9$\times$ to 1.1$\times$). In Table~\ref{tab:supp_s3dis} we show detailed per-category performance breakdown for our models and previous approaches.

\begin{table*}[h]
    \centering
    \tablestyle{3pt}{1.2}
    \resizebox{1.0\textwidth}{!}{
    \begin{tabular}{l|ccccccccccccc|cc}
    \toprule
Method & ceiling & floor & wall & beam & clmn & windw & door & chair & table & bkcase & sofa & board & clutter & mIOU & mAcc \\
\midrule
PointNet~\cite{qi2017pointnet}  & 88.80 & 97.33 & 69.80 & 0.05 & 3.92 & 46.26 & 10.76 & 52.61 & 58.93 & 40.28 & 5.85 & 26.38 & 33.22 & 41.09 & 48.98 \\
SegCloud~\cite{tchapmi2017segcloud} & 90.06 & 96.05 & 69.86 & 0.00 & 18.37 & 38.35 & 23.12 & 75.89 & 70.40 & 58.42 & 40.88 & 12.96 & 41.60 & 48.92 & 57.35 \\
TangentConv~\cite{tatarchenko2018tangent} & 90.47 & 97.66 & 73.99 & 0.0 & 20.66 & 38.98 & 31.34 & 77.49 & 69.43 & 57.27 & 38.54 & 48.78 & 39.79 & 52.8 & 60.7 \\
3D RNN~\cite{ye20183d} & 95.2 & 98.6 & 77.4 & 0.8 & 9.83 & 52.7 & 27.9 & 76.8 & 78.3 & 58.6 & 27.4 & 39.1 & 51.0 & 53.4 & 71.3 \\
PointCNN~\cite{li2018pointcnn} & 92.31 & 98.24 & 79.41 & 0.00 & 17.60 & 22.77 & 62.09 & 80.59 & 74.39 & 66.67 & 31.67 & 62.05 & 56.74 & 57.26 & 63.86 \\
SuperpointGraph~\cite{landrieu2018large} & 89.35 & 96.87 & 78.12 & 0.0 & 42.81 & 48.93 & 61.58 & 84.66 & 75.41 & 69.84 & 52.60 & 2.1 & 52.22 & 58.04 & 66.5 \\
MinkowskiNet20~\cite{choy20194d} & 91.55 & 98.49 & 84.99 & 0.8 & 26.47 & 46.18 & 55.82 & 88.99 & 80.52 & 71.74 & 48.29 & 62.98 & 57.72 & 62.60 & 69.62 \\
MinkowskiNet32~\cite{choy20194d} & 91.75 & 98.71 & 86.19 & 0.0 & 34.06 & 48.90 & 62.44 & 89.82 & 81.57 & \bf{74.88} & 47.21 & 74.44 & 58.57 & 65.35 & 71.71 \\
\midrule
PntContrast(Scratch) & 91.47 & 98.56 & 84.08 & 0.00 & 33.03 & 56.88 & 63.94 & 90.11 & 81.67 & 72.46 & 76.45 & 77.89 & 59.63 & 68.17 & 75.45 \\
PntContrast(Hardest-Ctr) & \bf{94.82} & \bf{98.72} & \bf{86.06} & 0.00 & 42.84 & \bf{58.00} & \bf{73.72} & \bf{91.73} & \bf{82.38} & 74.74 & 74.58 & 81.42 & \bf{62.66} & \bf{70.90} & \bf{77.00} \\
PntContrast(Pnt-InfoNCE) & 93.26 & 98.67 & 85.56 & 0.11 & \bf{45.90} & 54.41 & 67.87 & 91.56 & 80.09 & 74.66 & \bf{78.20} & \bf{81.49} & 62.32 & 70.32 & 76.94 \\
    \bottomrule
    \end{tabular}}
    \caption{\textbf{Stanford Area 5 Test (Fold 1).} Per-category IOU performance.}
    \label{tab:supp_s3dis}
\end{table*}

\section{Synthia4D Segmentation Experimental Details}
Here we provide training details for Synthia4D semantic segmentation task. As mentioned in the main paper, we only use 3D sparse convnet without any temporal aggregation mechanisms such as 4D kernels and temporal CRF. For all the PointContrast variants (Training from scratch, Hardest-contrastive Pretrained, and PointInfoNCE Pretrained) we use the same hyperparameter settings, and those are mostly identical the S3DIS experiments. Specifically we train the model with 8 V100 GPUs with data parallelism for 15,000 iterations. Batch size is 72. Batch normalization is applied independently on each GPU. We use SGD+momentum optimizer with an initial learning rate 0.8. We use Polynomial LR scheduler with a power factor of 0.9. Weight decay is 0.0001 and voxel size is 0.05 (5cm). We also use the same data augmentation techniques in \cite{choy20194d} in color space and point coordinate space. In Table~\ref{tab:supp_synthia} we show detailed per-category performance breakdown for our models and results reported in \cite{choy20194d}.
\begin{table*}[h]
    \centering
    \tablestyle{3pt}{1.2}
    \resizebox{1.0\textwidth}{!}{
    \begin{tabular}{l|cccccccccccc|cc}
    \toprule
        \small{Method}  & \small{Bldn} & \small{Road} & \small{Sdwlk} & \small{Fence} & \small{Vegittn} & \small{Pole} & \small{Car} & \small{T. Sign} & \small{Pedstrn} & \small{Bicycl} & \small{Lane} & \small{T. Light} & \small{mIoU} & \small{mAcc} \\
    \midrule
      MinkNet20 + TA \cite{choy20194d} & 88.096 & 97.790 & 78.329 & 87.088 & 96.540 & 97.486 & 94.203 & 78.831 & 92.489 & 0.000 & 46.407 & 67.071 & 77.03 & 89.198 \\
4D MinkNet32 + TS-CRF \cite{choy20194d} & 89.694 & 98.632 & 86.893 & 87.801 & 98.776 & 97.284 & 94.039 & 80.292 & 92.300 & 0.000 & 49.299 & 69.060 & 78.67 & 90.51 \\
    \midrule
PntContrast(Train from scratch) & 92.237 & 98.619 & 90.217 & 86.863 & 99.346 & 96.848 & 95.085 & 75.526 & 88.596 & 0.000 & 72.173 & 62.060 & 79.797 & 91.492 \\
PntContrast(Hardest-Contrastive) & \bf{92.518} & \bf{99.040} & 93.309 & 87.331 & 99.384 & 97.500 & \bf{96.174} & 81.627 & 92.007 & 0.000 & \bf{80.257} & 71.764 & 82.576 & 93.650 \\
PntContrast(PointInfoNCE) & 92.238 & 99.006 & \bf{93.993} & \bf{87.368} & \bf{99.657} & \bf{97.755} & 95.648 & \bf{83.446} & \bf{93.279} & 0.000 & 79.002 & \bf{76.364} & \bf{83.146} & \bf{93.707} \\
    \bottomrule
    \end{tabular}}
    \caption{\textbf{Synthia4D segmentation test results} Per-category IOU performance.}
    \label{tab:supp_synthia}
\end{table*}

\section{ScanNet Segmentation Experimental Details}
For ScanNet segmentation task, we train the model with 8 V100 GPUs with data parallelism for 15,000 iterations. Batch size is 48. We use SGD+momentum optimizer with an initial learning rate 0.8. We use Polynomial LR scheduler with a power factor of 0.9. Weight decay is 0.0001 and voxel size is 0.025 (2.5cm). We also use the same data augmentation techniques in \cite{choy20194d} in color space and point coordinate space. In Table~\ref{tab:supp_synthia} we show detailed per-category performance breakdown for our models and results reported in \cite{choy20194d}.
\begin{table*}[h]
    \centering
    \tablestyle{1.2pt}{1.2}
    \resizebox{1.0\textwidth}{!}{
    \begin{tabular}{l|cccccccccccccccccccc|cc}
    \toprule
        \small{Method}  & \small{bath} & \small{bed} & \small{bookshelf} & \small{cab} & \small{chair} & \small{counter} & \small{curtain} & \small{desk} & \small{door} & \small{floor} & \small{other} & \small{pic} & \small{ref} & \small{shower} & \small{sink} & \small{sofa} & \small{tab} & \small{toilet} & \small{wall} & \small{wind} & \small{mIoU} & \small{mAcc} \\
        \midrule
        Scratch &  84.866 & 96.412 & 64.271 & 80.928 & 90.910 & \bf{85.954} & 73.890 & 61.767 & 59.935 & 80.753 & 30.896 & \bf{68.571} & 62.109 & 75.252 & 54.603 & 67.236 & 90.084 & \bf{68.720} & 85.936 & 60.279 & 72.169 & 80.718 \\
Hardest-Ct & 85.773 & 96.442 & \bf{66.883} & \bf{81.525} & \bf{91.829} & 84.708 & 74.030 & 65.149 & 62.835 & \bf{82.645} & 32.113 & 66.451 & 64.408 & 77.507 & 53.049 & \bf{69.762} & 94.269 & 67.934 & \bf{88.601} & 60.190 & 73.305 & 81.025 \\
PntInfoNCE & \bf{86.540} & \bf{96.456} & 66.630 & 81.294 & 91.301 & 84.281 & \bf{77.393} & \bf{68.031} & \bf{65.168} & 81.600 & \bf{33.530} & 66.037 & \bf{67.639} & \bf{77.803} & \bf{56.853} & 69.398 & \bf{95.202} & 68.329 & 88.303 & \bf{60.924} & \bf{74.136} & \bf{81.623} \\
    \bottomrule
    \end{tabular}}
    \caption{\textbf{ScanNet segmentation results on val set} Per-category IOU performance.}
    \label{tab:supp_scannet}
\end{table*}

\section{ScanNet and SUN RGB-D Detection Details}
For the 3D object detection experiments, we mostly follow the configurations in VoteNet~\cite{voteNet} framework after switching in the SR-UNet backbone architecture. We train the model on 1 GPU, with batch size 64 for SUN RGB-D and 32 for ScanNet. Learning rate is 0.001 and we use Adam optimizer. The input points are subsampled before voxelization, we use 20000 points for SUN RGB-D and 40000 points for ScanNet. The voxel size is 2.5cm for ScanNet and 5cm for SUN RGB-D. In Table~\ref{tab:scannet:detection:supp} we show more results reported by previous methods. In Table~\ref{tab:supp_sunrgbd_det} and Table~\ref{tab:supp_scannet_det}, we show per-category AP performance for PointContrast models agains training from scratch results, under AP@0.5 metric. 
\begin{table*}[h]
    \centering
    \tablestyle{1.2pt}{1.2}
    \resizebox{1.0\textwidth}{!}{
    \begin{tabular}{l|cccccccccc|c}
    \toprule
Methods & bed\ \  & table & sofa & chair & toil & desk & dress & night & book & bath & mAP@0.5 \\
\midrule
Train from Scratch  & 47.8 & 19.6 & 48.1 & 54.6 & 60.0 & 6.3 & 15.8 & 27.3 & 5.4 & 32.1 & 31.7 \\
\midrule
Hardest-Contrastive & \bf{52.0} & \bf{20.1} & \bf{52.3} & \bf{55.8} & \bf{60.0} & \bf{7.5}& 14.7 & 36.8 & \bf{10.0} & 35.6 & 34.5 \\
PointInfoNCE       & 50.5 & 19.4 & 51.8 & 54.9 & 57.4 & \bf{7.5} & \bf{16.2} & \bf{37.0} & 5.9 & \bf{47.6} & \bf{34.8} \\
    \bottomrule
    \end{tabular}}
    \caption{\textbf{SUN RGB-D detection results} Per-category AP@0.5 performance.}
    \label{tab:supp_sunrgbd_det}
\end{table*}

\begin{table*}[h]
    \centering
    \tablestyle{1.2pt}{1.2}
    \resizebox{1.0\textwidth}{!}{
    \begin{tabular}{l|cccccccccccccccccc|c}
    \toprule
Methods & cabinet & bed & chair & sofa & table & door & wind & bkshlf & pic & cntr & desk & curtain & refrig & shower & toilet & sink & bath & garbage & mAP@0.5 \\
\midrule
Train from Scratch &  9.9 & 70.5 & 70.0 & 60.5 & 43.4 & \bf{21.8} & 10.5 & 33.3 & 0.8 & 15.4 & 33.3 & 26.6 & \bf{39.3} & 9.7 & 74.7 & 23.7 & 75.8 & 18.1 & 35.4 \\ 
\midrule
Hardest-Contrastive & 10.5 & 68.4 & \bf{75.6} & 59.1 & 43.1 & 19.6 & 9.6 & \bf{35.0} & \bf{2.1} & 15.6 & \bf{34.3} & \bf{32.8} & 37.8 & 13.6 & 76.9 & \bf{28.8} & \bf{82.4} & 25.8 & 37.3 \\
PointInfoNCE &       \bf{13.1} & \bf{74.7} & 75.4 & \bf{61.3} & \bf{44.8} & 19.8 & \bf{12.9} & 32.0 & 0.9 & \bf{21.9} & 31.9 & 27.0 & 32.6 & \bf{17.5} & \bf{87.4} & 23.2 & 80.8 & \bf{26.7} & \bf{38.0} \\
    \bottomrule
    \end{tabular}}
    \caption{\textbf{ScanNet detection results} Per-category AP@0.5 performance.}
    \label{tab:supp_scannet_det}
\end{table*}

\section{PointContrast vs FCGF for low- and high-level tasks} 
We take the best performing FCGF model released in \cite{choy2019fully} that achieves a high registration feature matching recall (FMR) of: 0.958. However, this model does not perform well for S3DIS segmentation. On the other hand, the PointContrast model that performs best for segmentation achieves a lower FMR when applied to the registration task. We conclude that low-level tasks and high-level tasks in 3D might require different design choices.
\begin{table*}[h]
\centering
\tablestyle{1.5pt}{1.1}
\begin{tabular}{c|c|c}
methods & Registration FMR & S3DIS mIoU \\
\hline
FCGF\cite{choy20194d} & \textbf{0.958} & 63.06 \\
\hline
PointContrast & 0.912 & \textbf{70.90} \\
\end{tabular}
\caption{\textbf{FCGF vs PointContrast.} FCGF achieves a much higher registration feature matching recall, while PointContrast achieves higher mIoU for segmentation.}
\label{tab:fcgfvspointcontrast}
\end{table*}

\begin{table*}[h]
\centering
\tablestyle{1.2pt}{1.1}
\scalebox{0.8}{\begin{tabular}{l|c|cc}
 methods & input & mAP@0.5 & mAP@0.25 \\
\shline
DSS~\cite{song2016deep,hou20183d} & Geo+RGB         & 6.8  & 15.2 \\
MRCNN 2D-3D~\cite{he2017mask,hou20183d} & Geo+RGB   & 10.5 & 17.3 \\
F-PointNet~\cite{qi2018frustum,hou20183d} & Geo+RGB & 10.8 & 19.8 \\
GSPN~\cite{yi2018gspn} & Geo+RGB                    & 17.7 & 30.6 \\ 
3D-SIS~\cite{hou20183d} & Geo+RGB \tiny{(5 Views)}  & 22.5 & 40.2 \\
VoteNet~\cite{voteNet} & Geo+Height                 & 33.5 & 58.6 \\
\hshline
Training from scratch & Geo                         & 35.4 & 56.7 \\
PointContrast(Hardest-Contrastive) & Geo            & 37.3 & \textbf{59.2}\\
PointContrast(PointInfoNCE) & Geo                   & \textbf{38.0} & 58.5 \\
\end{tabular}}
\caption{\textbf{3D object detection results on ScanNet dataset.} More methods in comparison. 
}
\label{tab:scannet:detection:supp}
\end{table*}

\end{document}